\documentclass[10pt,twocolumn,letterpaper]{article}

\usepackage{iccv}
\usepackage{times}
\usepackage{epsfig}
\usepackage{graphicx}
\usepackage{amsmath}
\usepackage{amssymb}
\usepackage{booktabs}
\usepackage{booktabs}
\usepackage{multirow}
\usepackage{threeparttable}
\usepackage{bbding}
\usepackage{makecell}
\usepackage[accsupp]{axessibility} 
	
\usepackage{color, colortbl}
\usepackage{booktabs}
\usepackage[dvipsnames]{xcolor}

\usepackage[pagebackref=true,breaklinks=true,letterpaper=true,colorlinks,bookmarks=false]{hyperref}

\iccvfinalcopy 

\def\iccvPaperID{5565} 

\ificcvfinal\fi

\begin{document}

\title{Decoupled DETR: Spatially Disentangling Localization and Classification for Improved End-to-End Object Detection}

\author{
Manyuan Zhang$^{1,2}$
\and
Guanglu Song$^2$
\and
Yu Liu$^2$
\and
Hongsheng Li$^{1,3,4}$
\and
{\tt\small $^1$ Multimedia Laboratory, The Chinese University of HongKong}
\and
{\tt\small $^2$SenseTime Research}
\and
{\tt\small $^3$Centre for Perceptual and Interactive Intelligence Limited}
\and
{\tt\small $^4$Shanghai AI Laboratory} \\
\small\texttt{zhangmanyuan@link.cuhk.edu.hk}
}

\maketitle
\ificcvfinal\thispagestyle{empty}\fi

\begin{abstract}
The introduction of DETR represents a new paradigm for object detection.
 However, its decoder conducts classification and box localization using shared queries and cross-attention layers, leading to suboptimal results. We observe that different regions of interest in the visual feature map are suitable for performing query classification and box localization tasks, even for the same object. Salient regions provide vital information for classification, while the boundaries around them are more favorable for box regression. Unfortunately, such spatial misalignment between these two tasks greatly hinders DETR's training.
 Therefore, in this work, we focus on decoupling localization and classification tasks in DETR. To achieve this, we introduce a new design scheme called spatially decoupled DETR (SD-DETR), which includes a task-aware query generation module and a disentangled feature learning process. 
We elaborately design the task-aware query initialization process and divide the cross-attention block in the decoder to allow the task-aware queries to match different visual regions.
 Meanwhile, we also observe that the prediction misalignment problem for high classification confidence and precise localization exists, so we propose an alignment loss to further guide the spatially decoupled DETR training.
Through extensive experiments, we demonstrate that our approach achieves a significant improvement in MSCOCO datasets compared to previous work. For instance, we improve the performance of Conditional DETR by 4.5 AP. By spatially disentangling the two tasks, our method overcomes the misalignment problem and greatly improves the performance of DETR for object detection.
\end{abstract}

\section{Introduction}
\label{sec:intro}

Object detection is a critical problem in computer vision, with traditional detectors relying on convolution to extract informative representations of the image, including single-stage and multi-stage detectors~\cite{ren2015faster,liu2016ssd,redmon2016you,bochkovskiy2020yolov4,lin2017feature}. In contrast, recent work, such as DETR~\cite{carion2020end}, breaks this paradigm. DETR consists of an encoder and a decoder, with the encoder extracting image features using self-attention and the decoder using these features to estimate object locations and categories in interaction with the object query. It is an end-to-end solution that does not require post-processing, such as non-maximum suppression (NMS).

\begin{figure}[t] 
\begin{center}
   \includegraphics[width=\linewidth]{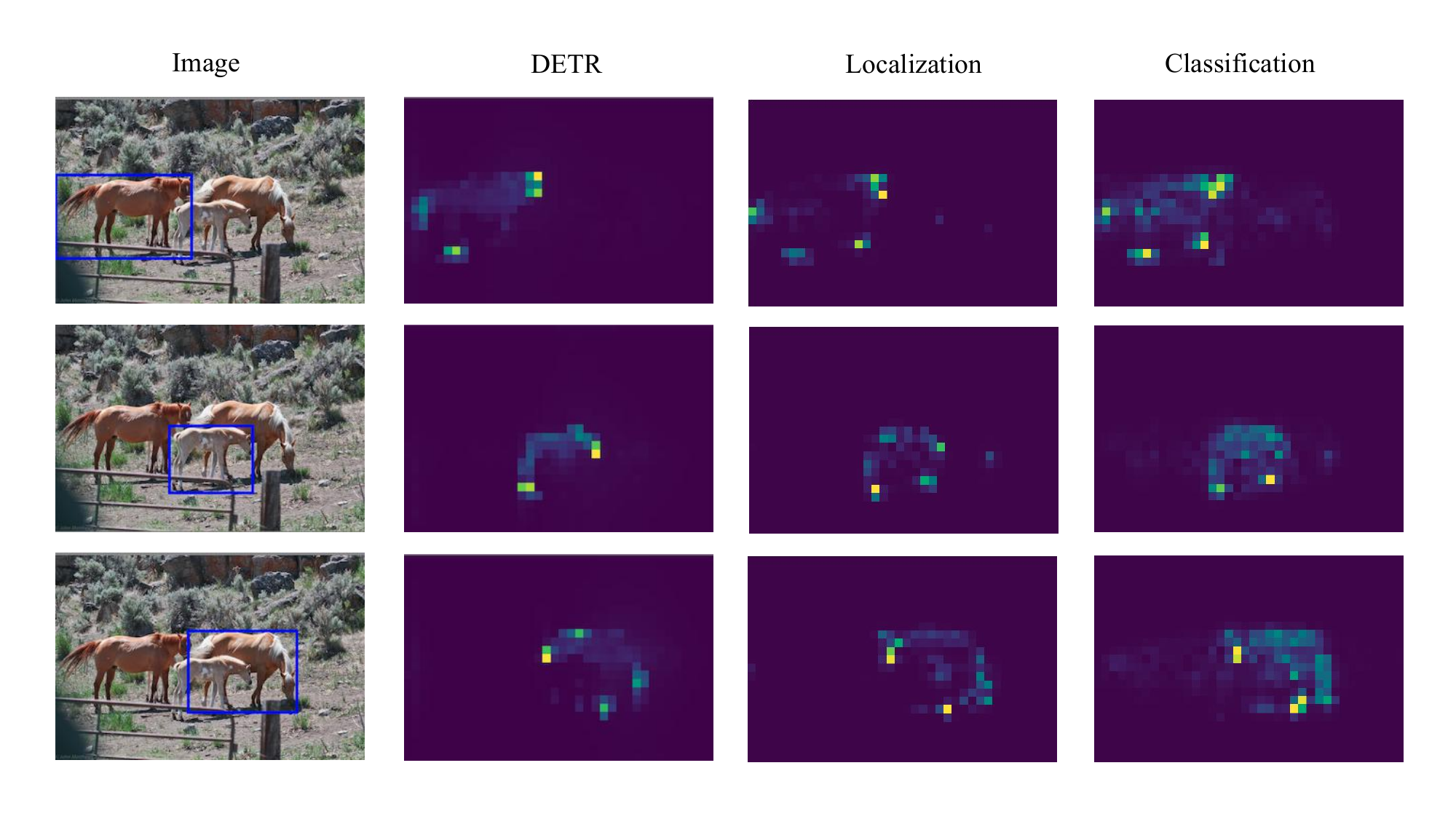}
\end{center}
   \caption{Visualization comparison of the cross-attention map for the original DETR and split decoder DETR. To explore the preference of the classification and localization branches for different regions, we make a copy of the transformer decoder so that the classification and localization branches can be decoupled entirely. There is a difference in the highlighted area between the two branches, and the localization branch is more focused on the object edges.
   }
\label{fig:intro}
\end{figure}

However, the lack of a good positional prior to matching the queries to the visual feature map has been observed to slow down DETR's convergence. As a result, subsequent work has focused on improving performance by developing various techniques for initializing object queries~\cite{liu2022dab,meng2021conditional,gao2021fast,wang2022anchor,zhang2022dino}. On the other hand, previous RCNN-based object detection work~\cite{jiang2018acquisition,wu2020rethinking,song2020revisiting} has shown that sharing the detection head for both classification and localization tasks may result in suboptimal performance, as the classification and localization branches may have conflicting learning targets and their actual regions of interest may not be well-aligned with each other. To address this issue, Double-Head R-CNN~\cite{wu2020rethinking} disentangles the detection head into two dedicated branches for classification and localization, respectively. While the disentanglement of the detection head can yield satisfactory performance, there persists a conflict between the two tasks due to the fact that the features fed into both branches originate from the same proposal through the use of ROI pooling. Additionally, as DETR relies heavily on attention mechanisms for information extraction, disentanglement methods used in previous RCNN-based detectors that rely on anchors or convolutional features cannot be directly transferred to the DETR-based detector.

Our study highlights the issue of misalignment between classification and localization tasks also exists in DETR, which has been ignored for a long time. To illustrate this problem, we conduct a pilot study to decouple the classification and localization branches entirely by creating two copies of the decoder. We then visualize the neuron activation maps for each of the two branches, as shown in Figure~\ref{fig:intro}. The second and third columns display the cross-attention maps for classification and localization branches, respectively. The highlighted activations of each branch are significantly different, indicating a  significant semantic misalignment. 
Further analysis shows that features from different positions within an object make varying contributions to classification and localization tasks. For instance, salient regions within an object provide vital information for classification, while the boundaries around objects are more favorable for box localization.

\begin{figure}[t!] 
\begin{center}
   \includegraphics[width=\linewidth]{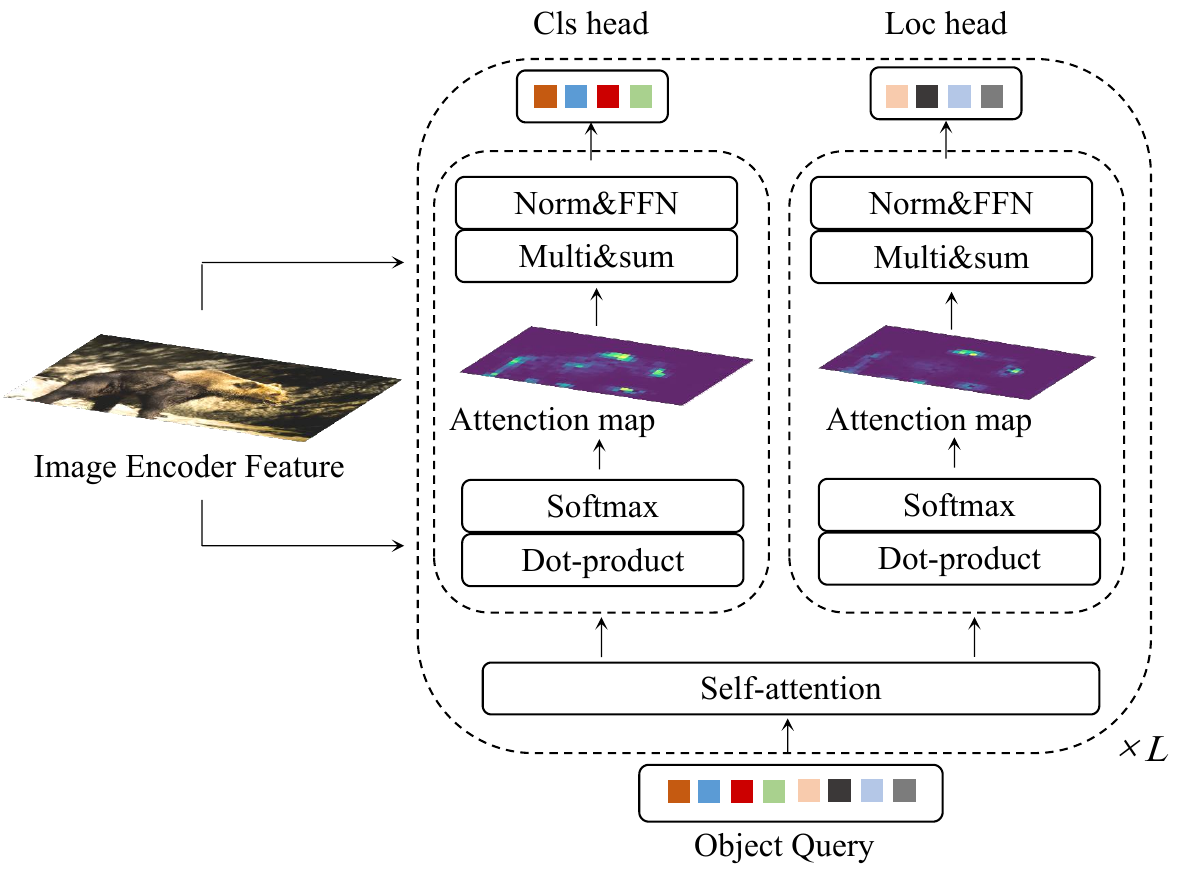}
\end{center}
   \caption{
The decoder architecture of our spatially decoupled DETR. We split the cross-attention block to allow the classification and localization branch can perform query matching with different regions of the visual feature map. And we keep the decoder self-attention block still shared so that queries from the two branches can better propagate information. 
   }
\label{fig:intro2}
\end{figure}

To take advantage of the observation, we propose a decoupled design scheme for the DETR decoder, as depicted in Figure~\ref{fig:intro2}. However, we do not naively adopt two totally isolated branches. Instead, we only split the cross-attention block in the decoder into two branches, allowing classification and localization to perform query matching with different regions of the visual feature map. Importantly, the two branches share the self-attention layers, enabling them to cooperate with each other in detecting the same objects. By decoupling cross-attention for classification and localization tasks, our proposed method achieves improved performance over existing DETR detectors.

Furthermore, we highlight the importance of query initialization in DETR's decoder for achieving good performance and convergence speed. In the original DETR, the input queries consist of a content query and a randomly initialized positional embedding. However, after decoupling the classification and localization branches, initializing the content and positional embeddings becomes critical. To address this, we introduce a task-aware query generation module that learns task-specific queries based on anchor boxes. We first find some discriminative points within the anchors boxes and then the content embedding initialization is sampled from the encoder's feature map for those discriminative points. While the positional embedding is generated using sinusoidal embedding of the offsets for those points.

We also observe that the misalignment problem between accurate classification and precise localization persists, resulting in high classification confidence with relatively low intersection-over-union (IoU) scores, or vice versa. Inspired by the aligned label assignment mechanism proposed in~\cite{feng2021tood,gao2021mutual,li2020generalized}, we further propose an alignment loss to guide the consistency between high classification confidence and precise localization in our Decoupled DETR learning framework.

We summarize our contributions as follows:
\begin{itemize}
\item We reveal the feature and prediction misalignment problem of the classification and localization branch in DETR, which significantly limits the performance of DETR-like detectors.
\item We disentangle the feature learning process for the classification and localization branches. We split the cross-attention in the decoder to allow the two branches to match different areas. We design task-aware query generation for better query initialization for the two branches.  We also propose an alignment loss to guide the consistency between high classification confidence and precise localization.
\item We integrate our structure into a wide range of variants of DETR, and a large number of experiments on MSCOCO demonstrate that our approach can lead to significant improvements.
\end{itemize}

\section{Related Work}

\subsection{Anchor-based Detector}

Deep learning have deeply changed the field of computer vision~\cite{zhang2020discriminability,zhang2022towards,liu2021switchable,shi2023videoflow,shi2023flowformer++,liu2021fuseformer,liu2019towards,huang2022flowformer}, and one widely studied area is object detection. In the past, a significant number of object detection methods have relied on anchors, such as the Faster R-CNN~\cite{ren2015faster,cai2018cascade} and YOLO~\cite{liu2016ssd,bochkovskiy2020yolov4,redmon2016you,lin2017focal} series, which predict the offsets between the detected object and predefined anchors. Other approaches, such as CornerNet~\cite{law2018cornernet}, CenterNet~\cite{duan2019centernet}, Scale-aware Detectors~\cite{li2019scale}, and FCOS~\cite{zhu2019feature}, utilize anchor points to characterize the prior and regress the distance between the object and the points. The anchor mechanism provides a useful assumption for object detection.

\subsection{DETR and its Variants}

The proposed DETR~\cite{carion2020end} provides a novel paradigm for object detection that relies on an encoder for further extraction of image content and a decoder for information matching between queries and encoder features to obtain object category and position information. Unlike traditional detectors, DETR eliminates the need for non-maximum suppression (NMS). However, the original DETR suffers from several issues, such as high computational complexity and slow convergence speed.

To tackle the issue of computational complexity, PNP-DETR~\cite{wang2021pnp} employs a poll-and-pool strategy to combine redundant tokens while retaining only valuable ones. Deformable-DETR~\cite{zhu2020deformable} uses the deformable mechanism to dynamically select tokens for multi-head attention calculation, significantly reducing the computational effort and making the complexity independent of the number of tokens.

As for the slow convergence issue, the original DETR typically requires 500 epochs to converge. Many works attribute this to completely random object query initialization in the decoder. SMCA~\cite{gao2021fast} introduces a Gaussian mechanism to constrain global cross-attentions to focus more on specific regions, reducing the difficulty of matching. Conditional DETR~\cite{meng2021conditional} decouples content from positional matching, allowing the model to search the extremity. Anchor-DETR~\cite{wang2022anchor} introduces the anchor mechanism, where an anchor corresponds to a query, making the query initialization interpretable. DAB-DETR~\cite{liu2022dab} modulates the positional attention map using box width and height information based on the anchor box, making the matching prior more specific. DN-DETR~\cite{li2022dn}, Group DETR~\cite{chen2022group}, and Hybrid DETR~\cite{jia2022detrs} focus on the inefficiency of one-to-one matching on the Hungarian loss and transform it into one-to-many matching to accelerate convergence.

A former work by He et al.~\cite{he2022destr} also addresses the feature misalignment problem of the classification and localization branch in DETR by splitting the cross-attention layer. However, their approach overlooks the importance of proper query embedding initialization and the misalignment problem for high classification confidence and precise localization of detection results.

\begin{figure*}[t!] 
\begin{center}
   \includegraphics[width=1.0\linewidth]{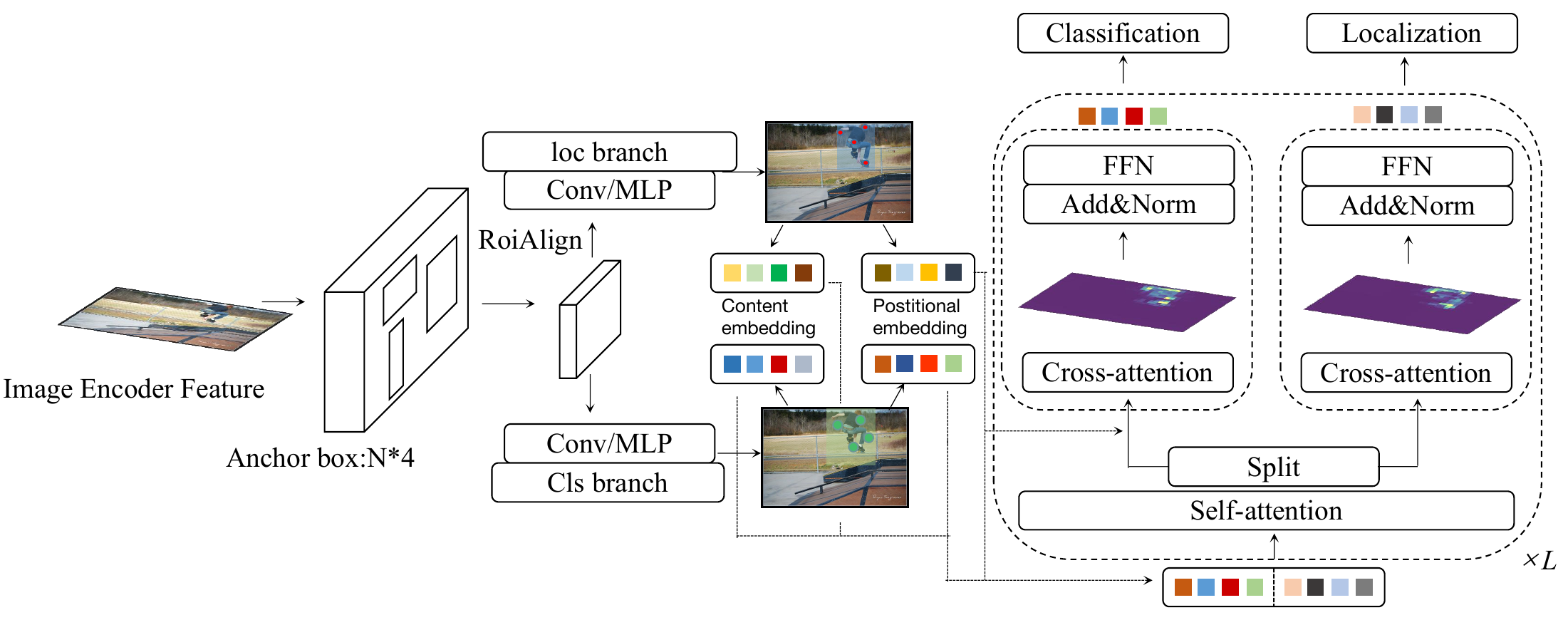}
\end{center}
\caption{ The pipeline of the proposed spatially decoupled DETR architecture, which addresses the issue of feature misalignment between the two branches of the cross-attention block in the decoder. Specifically, the cross-attention block is split, while the self-attention block is shared to enable information propagation. To enhance task-disentangled feature learning, we find the most discriminative points and use them to initiate task-aware content and positional embedding for object queries of the two branches.}
\label{fig:architecture}
\end{figure*}

\subsection{Feature Decoupling Methods}

The misalignment between the classification and localization branches has been extensively investigated in object detection during the convolution era. For example, IoU-Net~\cite{jiang2018acquisition} found that the feature generating a high classification score typically predicts a coarse bounding box. To address this, they introduced an additional head to predict the Intersection over Union (IoU) as the localization confidence and then combined this with the classification confidence to obtain the final classification score. Double-Head R-CNN~\cite{wu2020rethinking} disentangles the sibling head into two separate branches for classification and localization. Another approach, TSD~\cite{song2020revisiting}, spatially disentangles the gradient flows for classification and localization. While some label assignment methods such as TOOD~\cite{feng2021tood} and MuSu~\cite{gao2021mutual} propose a new anchor alignment metric integrated into the sample assignment and loss functions to dynamically guide the consistency between high classification confidence and precise localization. While those methods have been successful in the convolution era, their efficacy in DETR requires further validation.

\section{Method}

In this section, we first revisit the original design scheme of DEtection TRansformer (DETR)~\cite{carion2020end} and then describe our proposed spatially decoupled DETR (SD-DETR).
The sub-modules will be introduced in Section~\ref{sec321} and Section~\ref{sec322}. Then a novel alignment loss that further guides the consistency of high classification confidence and precise localization will be proposed in Section~\ref{sec323}. Finally, we delve into the inherent problem in the original query and decoder fully shared by classification and localization and demonstrate the advantage of our spatially decoupled DETR.

\subsection{Revisit the General DETR Pipeline}

DETR is a flexible end-to-end detector that views object detection as a set prediction problem. It pre-defines multiple queries and introduces a one-to-one set matching scheme based on a transformer encoder-decoder architecture. Each ground truth is assigned to a specific query as the supervised target for classification and localization. Specifically, given an input image $I$, its visual feature $\operatorname{F}$ can be generated by the backbone and transformer encoder. Let $Q=\{q_0, \dots, q_n\}$ denotes the pre-defined data-independent queries. The output query embeddings can be generated via:
\begin{equation}
\label{eq1}
\hat{Q} = \{\operatorname{self-att}(Q), \operatorname{cross-att}(Q,\operatorname{F}), \operatorname{FFN}(Q)\}_{\times L},
\end{equation}
where $\operatorname{self-att}(\cdot)$ are the self-attention  that propagates information between queries $q_i$, $\operatorname{cross-att}(\cdot)$ is the cross-attention  that absorbs knowledge from $\operatorname{F}$ to support classification and localization, and $\operatorname{FFN}(\cdot)$ is the feedforward network. These operations are serially stacked for $L$ times to establish DETR's decoder. $Q$ is updated after being forwarded in each operation. Then, DETR applies task-specific prediction heads on $\hat{Q}$ to generate a set of predictions $\hat{P} = \{p_0,\dots,p_n\}$, which can be formulated as:
\begin{equation}
p_i = (p^{cls}_i, p^{loc}_i) = (\mathrm{F}_\mathrm{cls}(\hat{q_i}), \mathrm{F}_\mathrm{loc}(\hat{q_i})),
\end{equation}
where $\hat{q}_i \in \hat{Q}$, $p^{*}_{i}$ is the predictions for object classification or localization, and $\mathrm{F_{*}}$ indicates the two heads. Finally, DETR adopts the one-to-one bipartite matching to assign ground truths to  $\hat{P}$. In the original DETR training, all operations in Eq. (\ref{eq1}) use a shared-weight decoder.

\subsection{Spatially Decoupled DETR} 
\label{sec32}

As we analyzed above, the inherent conflict caused by the shared queries in different tasks, i.e., classification and localization, and the shared cross-attention operation in different queries greatly limits the performance of DETR-based detectors. For one instance, the features in some salient areas may have rich information for classification, while these around the boundary may be good at bounding.
The fully shared paradigm in DETR impedes it from learning better task-specific features to further improve its performance.

For this potential problem, we introduce the spatially decoupled DETR to alleviate this conflict by disentangling the tasks from two aspects, disentangled feature learning (DFL)  and task-aware query generation.
In DFL, Eq.(\ref{eq1}) is adjusted by:
\begin{small}
\begin{equation}
\begin{split}
\hat{Q}=\left\{ \operatorname{self-att(cat(Q_{cls}, Q_{loc}))}, \right. \\
\left. \begin{matrix}
\operatorname{cross-att_{cls}(Q_{cls})}, \operatorname{FFN_{cls}(Q_{cls})} \\
\operatorname{cross-att_{loc}(Q_{loc})}, \operatorname{FFN_{loc}(Q_{loc})}
\end{matrix} \right\}_{\times L}
\end{split}
\end{equation}

\end{small}
where $Q_{cls}$ and $Q_{loc}$ are the classification-aware and localization-aware queries generated by the task-aware query generation module.
The cross-attention module and FFN module are not shared between $Q_{cls}$ and $Q_{loc}$.
This generation process is formulated as follows:
\begin{equation}
\label{eq4}
\operatorname{Q_{cls} = G_{cls}(F, R_{box})},
\end{equation}
where $\operatorname{R_{box}}$ is a series of  anchor boxes. We use the mini-detector module proposed in ~\cite{he2022destr,zhu2020deformable} to initialize those anchor boxes.  $G_{cls}$ is the task-aware query generation process which will be introduced in \ref{sec322}

By disentangling the queries and feature learning in cross-attention and FFN, our spatially decoupled DETR can learn the better task-aware feature representation adaptively.
It's applicable to most existing DETR-based detectors whilst introducing few overheads. The overall pipeline of our spatially decoupled DETR can be seen in Figure~\ref{fig:architecture}.

\subsubsection{ Disentangled Feature Learning}
\label{sec321}

To fully utilize the network's capability, we need to design a disentangled feature learning architecture for these two sub-tasks. The simplest approach is to directly split the transformer decoder, but this results in a suboptimal design due to a lack of information propagation between the two branches and a significant increase in model parameters.

In this work, we keep the self-attention block 
 shared and split the cross-attention block in the decoder. The two branches share the self-attention block to enhance the information propagated between them.
Obtained the task-aware object query initialization, we concatenate the
classification query and localization query for the self-attention block.
So the output for the self-attention block will be $O_{self} \in \mathbb{R}^{2 C}$. Then we split the $O_{self} $ into two parts for the two branches, each for $C$-dimensional vector. Those two features will conduct cross-attention separately with the image feature output from the encoder. Each branch will search its matching interest area and distill relevant features, avoiding feature misalignment between those two branches.

When conducting the cross-attention, we concatenate the content query output from the split self-attention and positional embedding initialized based on the task-aware anchor prior to forming the object query for conditional matching ~\cite{meng2021conditional,liu2022dab}. The cross-attention output $O_{cross}^{\text {cls }} \in \mathbb{R}^{C}$ and $O_{cross}^{\text {loc }} \in \mathbb{R}^{ C}$ are passed to normalization, FFN, and residual link layer and then to the next decoder stage. Then the classification head and box regression head will take the final stage disentangled feature output as input, respectively, and finally obtain the object's category and location.

\subsubsection{Task-Aware Query Generation}
\label{sec322}

In the previous section, we divided the cross-attention block in the decoder for disentangled feature learning. In this section, we focus on generating task-aware queries to enable each branch's cross-attention to concentrate on their respective regions of interest and eliminate feature conflicts. We introduce the task-aware query generation module. Building upon the mini-detector proposed in~\cite{zhu2020deformable,he2022destr}, we begin by obtaining a set of anchor boxes $\operatorname{{R}_{box}}$. Next, we modify the semantic-aligned matching module proposed in~\cite{zhang2022accelerating}. Specifically, we employ ROIAlign to extract region-level features $\operatorname{{F}_{R}} \in \mathbb{R}^{N \times 7 \times 7 \times d}$ from the encoded features $\operatorname{F}$ for the region  of the anchor boxes $\operatorname{{R}_{box}}$.

To effectively capture object features within the anchor boxes, we select the most discriminative points to generate object content embeddings. As illustrated in Figure~\ref{fig:architecture}, we utilize a ConvNet and an MLP to obtain the coordinates $\operatorname{{R}_{SP}} \in \mathbb{R}^{N \times M \times 2}$ for these points within each region after obtaining region features ${F}_{\mathrm{R}}$ through RoIAlign.

\begin{equation}
\operatorname{{R}_{SP}}=\operatorname{MLP}\left(\operatorname{ConvNet}\left(\operatorname{{F}_{R}}\right)\right)
\end{equation}

These discriminative points are crucial for object recognition and localization. We employ bilinear interpolation to obtain their features.
We then calculate the average feature and offsets from these discriminative points for each branch. Subsequently, we use these average features to update the query content embedding. Meanwhile, we employ the positional encoding function PE to generate positional embeddings for the average offsets, which are used to update the positional embedding for the learnable query.

\subsubsection{Task  Alignment Learning}
\label{sec323}

In the previous two sections, we discuss the decoupling of the classification and localization branches in DETR. However, the misalignment between accurate classification and precise localization can significantly hinder the effectiveness of learning when generating predictions from object queries. This misalignment refers to situations where a query yields a high confidence classification but relatively low intersection-over-union (IoU) scores, or vice versa. Inspired by previous label assignment work \cite{feng2021tood,gao2021mutual,li2020generalized}, we make modifications to the loss function of DETR. Our goal is to ensure that both high classification scores and precise localization are achieved simultaneously. To accomplish this, we measure the task alignment based on a high-order combination of the classification score and the IoU. Specifically, we  design the following metric to calculate the alignment for each query:

\begin{equation}
t=s^\alpha \times u^\beta
\end{equation}

Here, $s$ represents the classification score, and $u$ denotes the IoU value. The parameters $\alpha$ and $\beta$ are utilized to control the relative impact of the two tasks in the alignment metric.  We then use $t$ to replace the binary label of positive samples during training, which encourages the learning process to dynamically prioritize high-quality queries. The Binary Cross Entropy (BCE)  for the classification task can be rewritten as:

\begin{equation}
\mathcal{L}_{cls}=\sum_{i=1}^{N_{\text{pos}}} |t_i-s_i|^\gamma BCE(s_i, t_i)+\sum_{j=1}^{N_{\text{neg}}} s_j^\gamma BCE(s_j, 0),
\end{equation}

Here, $N_{\text{pos}}$ and $N_{\text{neg}}$ represent the number of positive and negative samples, respectively, and $\gamma$ is the focusing parameter. To further increase the matching efficiency, we repeat the positive label for several times to provide a richer positive supervised signal.

\begin{figure}[t] 
\begin{center}
   \includegraphics[width=0.988\linewidth]{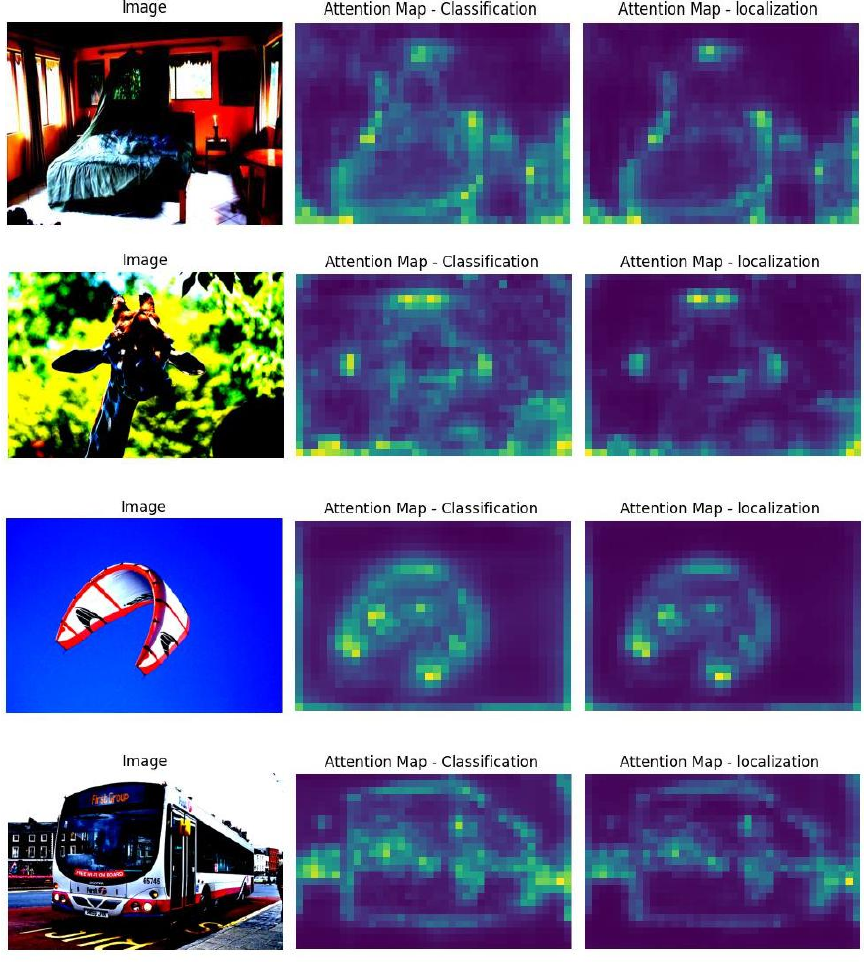}
\end{center}
   \caption{
Visualization of cross-attention maps for the classification branch and the localization branch for our spatially decoupled DETR. We randomly sampled four images from the COCO datasets, and the maps are softmax normalized over the dot products. We average the attention map of 8 heads. The high activation area differs for the two branches. 
   }
\label{fig:visualization}
\end{figure}

\subsection{Discussion}

Figure~\ref{fig:visualization} presents the cross-attention maps for the classification and localization branches. These maps are obtained by applying softmax normalization over the dot products and visualizing the average map of 8 cross-attention heads, for example, images randomly sampled from the COCO dataset. The spatial attention maps demonstrate that the classification and localization tasks have different high activation areas, which supports our hypothesis that features from different object positions contribute differently to these tasks. The localization branch tends to have a higher activation towards object edges, while the classification branch pays more attention to the overall object, especially the salient region. By decoupling the two branches, each one has more flexibility to capture its unique information.

\begin{table*}[t]
\begin{center}
\centering
\setlength{\tabcolsep}{12pt}
\resizebox{1\textwidth}{!}{
\small
\begin{tabular}[t]{l|c|c|cccccc}

\toprule[1.6666pt]
Method & multi-scale & Epochs &  AP & AP$_{\rm 0.5}$ & AP$_{\rm 0.75}$ & AP$_{\rm S}$ & AP$_{\rm M}$ & AP$_{\rm L}$ \\

\midrule[1.2333pt]

\multicolumn{9}{l}{\textit{Baseline methods trained for long to addresss:}} \\

\midrule[0.6666pt]

Faster-RCNN-R50-DC5~\cite{ren2015faster} & & 108   & 41.1 & 61.4 & 44.3 & 22.9 & 45.9 & 55.0 \\

Faster-RCNN-FPN-R50~\cite{ren2015faster,lin2017feature} & $\checkmark$ & 108  & 42.0 & 62.1 & 45.5 & 26.6 & 45.4 & 53.4 \\

DETR-R50~\cite{carion2020end} & & 150  & 42.0 & 62.4 & 44.2 & 20.5 & 45.8 & 61.1 \\

DETR-R50-DC5~\cite{carion2020end} & & 150  & 43.3 & 63.1 & 45.9 & 22.5 & 47.3 & 61.1 \\

\midrule[0.6666pt]

\multicolumn{9}{l}{\textit{Comparison of  with other detectors under 12 epochs training schemes:}} \\

\midrule[0.6666pt]

Faster-RCNN-R50~\cite{ren2015faster} & & 12  & 35.7 & 56.1 & 38.0 & 19.2 & 40.9 & 48.7 \\

DETR-R50~\cite{carion2020end}  & & 12  & 22.3 & 39.5 & 22.2 & 6.6 & 22.8 & 36.6 \\

Deformable-DETR-R50~\cite{zhu2020deformable} & & 12 &   31.8 & 51.4 & 33.5 & 15.0 & 35.7 & 44.7 \\

Conditional-DETR-R50~\cite{meng2021conditional} & & 12 &   32.2 & 52.1 & 33.4 & 13.9 & 34.5 & 48.7 \\

SMCA-DETR-R50~\cite{gao2021fast} & & 12  & 31.6 & 51.7 & 33.1 & 14.1 & 34.4 & 46.5 \\

SAM-DETR-R50~\cite{zhang2022accelerating}  & & 12  & 33.1 & 54.2 & 33.7 & 13.9 & 36.5 & 51.7 \\

DESTR-R50~\cite{he2022destr}  & & 12 & 35.9 & 56.8 & 37.2 & 16.2 & 39.2 & 53.1 \\

DAB-DETR-R50~\cite{liu2022dab}  & & 12  & 34.9 & 55.3 & 36.5 & 16.4 & 38.1 & 51.3 \\

\rowcolor[gray]{.9}
\textbf{SD-DETR-R50 } & & 12 &  \textbf{39.5} & \textbf{59.3} & \textbf{41.7} & \textbf{19.3} &\textbf{ 43.2} & \textbf{57.0 }\\

\midrule[0.6666pt]

Faster-RCNN-R50-DC5~\cite{ren2015faster} & & 12 &  37.3 & 58.8 & 39.7 & 20.1 & 41.7 & 50.0 \\

DETR-R50-DC5~\cite{carion2020end}  & & 12 &  25.9 & 44.4 & 26.0 & 7.9 & 27.1 & 41.4 \\

Deformable-DETR-R50-DC5~\cite{zhu2020deformable} & & 12  & 34.9 & 54.3 & 37.6 & 19.0 & 38.9 & 47.5 \\

Conditional-DETR-R50-DC5~\cite{meng2021conditional} &  & 12  & 35.9 & 55.8 & 38.2 & 17.8 & 38.8 & 52.0 \\

SMCA-DETR-R50-DC5~\cite{gao2021fast} &  & 12  & 32.5 & 52.8 & 33.9 & 14.2 & 35.4 & 48.1 \\

SAM-DETR-R50-DC5~\cite{zhang2022accelerating}  & & 12  & 38.3 & 59.1 & 40.1 & 21.0 & 41.8 & 55.2 \\

DESTR-R50-DC5~\cite{he2022destr}  & & 12  & 37.2 & 57.5 & 39.2 & 18.9 & 40.5 & 53.2 \\

DAB-DETR-R50-DC5~\cite{liu2022dab}  & & 12  & 37.7 & 58.0 & 40.1 & 19.6 & 41.5 & 52.1 \\

\rowcolor[gray]{.9}
\textbf{SD-DETR-R50 } & & 12  & \textbf{40.6} & \textbf{59.9} & \textbf{43.1} & \textbf{21.7} & \textbf{44.2} & \textbf{56.5} \\

\midrule[0.6666pt]

\multicolumn{9}{l}{\textit{Comparison of  with other detectors under 50 epochs training schemes:}} \\

\midrule[0.6666pt]

Faster-RCNN-R50~\cite{ren2015faster} & & 36  & 38.4 & 58.7 & 41.3 & 20.7 & 42.7 & 53.1 \\

DETR-R50~\cite{carion2020end}  & & 50  & 34.9 & 55.5 & 36.0 & 14.4 & 37.2 & 54.5 \\

Deformable-DETR-R50~\cite{zhu2020deformable} & & 50  & 39.4 & 59.6 & 42.3 & 20.6 & 43.0 & 55.5 \\

Conditional-DETR-R50~\cite{meng2021conditional} & & 50 & 41.0 & 61.8 & 43.3 & 20.8 & 44.6 & 59.2 \\

SAM-DETR-R50~\cite{zhang2022accelerating} & & 50  & 39.8 & 61.8 & 41.6 & 20.5 & 43.4 & 59.6 \\
DESTR-R50~\cite{he2022destr} & & 50  & 43.6 & 64.7  & 46.5 & 23.6 & 47.5 & 62.1 \\

DAB-DETR-R50~\cite{liu2022dab} & & 50  & 42.2 & 63.1 & 44.7 & 21.5 & 45.7 & 60.3 \\

\rowcolor[gray]{.9}
\textbf{SD-DETR-R50} & & 50  & \textbf{45.5} & \textbf{65.4 } & \textbf{48.5} & \textbf{25.6} &\textbf{ 49.9} & \textbf{64.2} \\

\midrule[0.6666pt]

Deformable-DETR-R50~\cite{zhu2020deformable} & $\checkmark$ & 50  & 43.8 & 62.6 & 47.7 & 26.4 & 47.1 & 58.0 \\

SMCA-DETR-R50~\cite{gao2021fast} & $\checkmark$ & 50  & 43.7 & 63.6 & 47.2 & 24.2 & 47.0 & 60.4 \\

DAB-Deformable-DETR-R50~\cite{liu2022dab} & $\checkmark$ & 50  & 46.8 & 66.0 & 50.4 & 29.1 & 49.8&  62.3 \\

\midrule[0.6666pt]

Faster-RCNN-R50-DC5~\cite{ren2015faster} & & 36  & 39.0 & 60.5 & 42.3 & 21.4 & 43.5 & 52.5 \\

DETR-R50-DC5~\cite{carion2020end}  & & 50  & 36.7 & 57.6 & 38.2 & 15.4 & 39.8 & 56.3 \\

Deformable-DETR-R50-DC5~\cite{zhu2020deformable} & & 50  & 41.5 & 61.8 & 44.9 & 24.1 & 45.3 & 56.0 \\

Conditional-DETR-R50-DC5~\cite{meng2021conditional} & & 50 & 43.8 & 64.4 & 46.7 & 24.0 & 47.6 & 60.7 \\

SAM-DETR-R50-DC5\cite{zhang2022accelerating} & & 50 & 43.3 & 64.4 & 46.2 & 25.1 & 46.9 & 61.0 \\

DESTR-R50-DC5~\cite{he2022destr} & & 50  &45.3 & 65.7 & 48.3& 27.3 &  48.8 &  62.4  \\
DAB-DETR-R50-DC5~\cite{liu2022dab} & & 50  &44.5 & 65.1 & 47.7 & 25.3 & 48.2 & 62.3  \\
\rowcolor[gray]{.9}
\textbf{SD-DETR-R50-DC5} & & 50  &\textbf{47.0} & \textbf{66.5} & \textbf{50.2} & \textbf{28.9} & \textbf{51.0} & \textbf{64.4}  \\

\bottomrule[1.6666pt]
\end{tabular}}
\end{center}
\caption{Comparison of the proposed SD-DETR, other DETR-like detectors, and Faster R-CNN on MSCOCO validation set.  We report the results with multiple backbones. Some method results are reported by~\cite{zhang2022accelerating}.}
    \vspace{-0.4cm}
\label{tab:main result}
\end{table*}

\section{Experiments}

\subsection{Experimental Setup}

\paragraph{Dataset}

All experiments are conducted on the COCO 2017 dataset \cite{lin2014microsoft}, which consists of 117k training examples and 5k validation images.

\paragraph{Training}

We follow the standard training protocol for DETR \cite{jia2022detrs}. We use ResNet \cite{he2016deep} as backbones from the TORCHVISION ImageNet-pretrained model zoo. The batch norm layers are fixed, and the transformer parameters are initialized with the Xavier initialization scheme.

We use the AdamW optimizer \cite{loshchilov2017decoupled} and train for 50 epochs. The weight decay is set to $10^{-4}$. The learning rate for the backbone and transformer is $10^{-5}$ and $10^{-4}$, respectively. The learning rate is dropped by ten after 40 epochs. We use a dropout rate of 0.1 for the transformer. We keep the number of multi-head to 8 and the attention channel to 256. The default query number is 300. To ensure a fair comparison with other works, we may change the training settings. For the architecture, we use six encoder layers and six decoder layers. We use bipartite matching via the Hungarian algorithm when calculating the loss function. We repeat the positive samples twice. For the task alignment loss, the $\alpha$ is set to 0.25 and  $\beta$ 0.75 respectively. The focusing parameter is 2, following~\cite{lin2017focal}. For the box regression loss, we apply the L1 and generalized IoU loss. We use the same data augmentation as DETR. We resize the input image to the short side within the range [480, 800] and the long side to 1333 pixels. We also conduct a random crop with a probability of 0.5.

\paragraph{Evaluation}

We follow the standard COCO evaluation and report the average precision (AP) at 0.50, 0.75 for small, medium, and large objects.
 
\subsection{Results}

Table~\ref{tab:main result} presents our main results on the COCO 2017~\cite{lin2014microsoft} validation set. We compared our proposed spatially decoupled DETR with various state-of-the-art methods including DETR~\cite{carion2020end}, Faster RCNN~\cite{ren2015faster}, Anchor DETR~\cite{wang2022anchor}, SMCA~\cite{gao2021fast}, Deformable DETR~\cite{zhu2020deformable}, Conditional DETR~\cite{meng2021conditional}, DAB-DETR~\cite{liu2022dab}, DESTR~\cite{he2022destr}. Our SD-DETR outperformed all previous methods by a significant margin, achieving a 4.5 AP improvement compared to Conditional DETR.  In comparison to DESTR, our method achieves a performance gain of 1.9 AP.

\begin{figure*}[t] 
\begin{center}
   \includegraphics[width=0.85\linewidth]{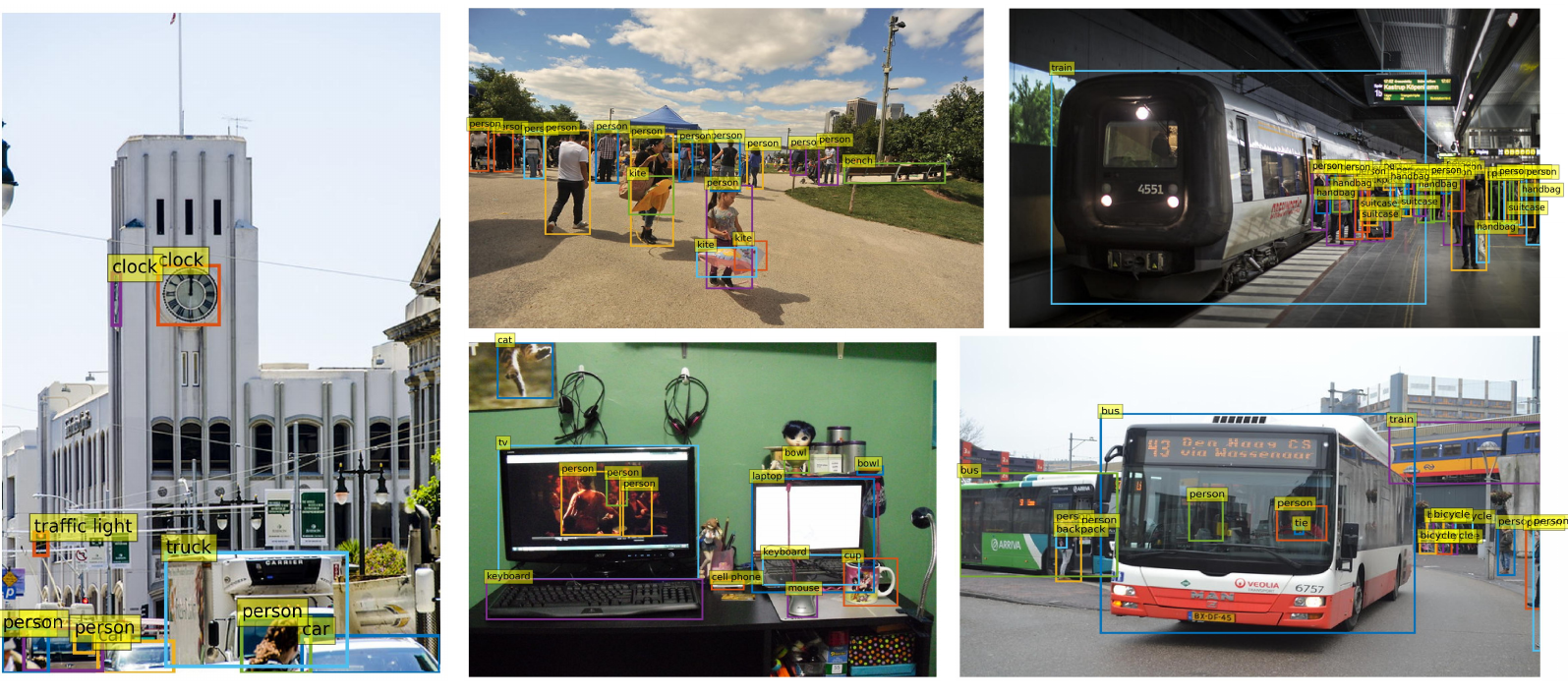}
\end{center}
   \caption{
Visualization of detection results for our SD-DETR. By addressing the issue of misalignment between the classification and localization branches, we have achieved a more robust detection performance, particularly in complex scenarios.
   }
\label{fig:vis_results}
\end{figure*}

We also tested a stronger backbone, R50-DC5, and our proposed spatially decoupled DETR consistently improved the performance of the original DETR and its variants under various settings. We outperform Conditional DETR for 3.2 AP and DAB-DETR for 2.5 AP.
These results demonstrate that decoupling the classification and localization branches in the DETR decoder eliminates both feature and prediction misalignment, enabling accurate classification and localization at different locations. In Figure~\ref{fig:vis_results}, we show some detection results of our SD-DETR, which show remarkable performance even in highly complex scenarios.

\vspace{-1mm}

\subsection{Ablations}

We conducted a comprehensive analysis of each component in our spatially decoupled DETR and assessed its impact on the final results, as shown in Table~\ref{tab:table2}. The table demonstrates that the performance of our spatially decoupled DETR improves gradually with the incorporation of different modules. In the top row of the table, we show the results of our baseline, Conditional DETR~\cite{meng2021conditional} with mini-detector~\cite{he2022destr,zhu2020deformable}. By employing disentangled feature learning, which decouples the cross-attention layer while maintaining shared self-attention, we achieve a 1.4 AP performance gain, highlighting the necessity of decoupling. Furthermore, through task-aware query generation, which generates more informative content and positional embeddings initialization for each branch, we further enhance the performance to 43.6. Lastly, by improving the loss function of the original DETR to address prediction misalignment issues in high-confidence classification and precise localization, we significantly boost the performance to 45.5. These three contributions of our work are propagated and all are aimed at resolving the misalignment problem in DETR's classification and localization.

\vspace{-4mm}
\paragraph{Comparison of fully split decoder}

In this section, we investigate the impact of different decoupling structures in the decoder. The most straightforward decoupling structure is a direct copy of the decoder, with the classification branch entirely decoupled from the localization branch. However, as shown in Table~\ref{tab:table3}, this structure only results in a 0.5 performance gain. Full decoupling ignores the information propagation between the two branches. Therefore, in our design, we split the cross-attention and share self-attention, which preserves information propagated between queries of different branches, resulting in a 0.9 gain and introducing fewer extra parameters.

\begin{table}[]
\centering
\resizebox{0.53\linewidth}{!}{%
\begin{tabular}{llll}
\bottomrule
DFL & TAQ & TAL & AP   \\ \hline
    &     &     & 41.7 \\
 \checkmark   &     &     & 43.1 \\
  \checkmark  &  \checkmark   &     & 43.6 \\
  \rowcolor[gray]{.9}
 \checkmark   &  \checkmark   & \checkmark    & \textbf{45.5} \\ \hline
\end{tabular}%
}
\caption{Ablation study for different components in our spatially decoupled DETR. The results are reported on the MSCOCO validation set. We gradually add new components. DFL refers to disentangled feature learning, which decouples the cross-attention layer. While TQG means task-aware query generation.  TAL means task alignment learning, which optimizes the loss function.    }
\label{tab:table2}
\end{table}

\begin{table}[t]
\centering
\def\arraystretch{1.13}
\resizebox{0.54\linewidth}{!}{
\begin{tabular}{c|c}
\toprule
Model         &        AP   \\ \midrule
Baseline                      & 41.7 \\
Split decoder              & 42.2 \\
\rowcolor[gray]{.9}
Split cross-attention      & \textbf{43.1} \\ \bottomrule
\end{tabular}}
\caption{Performance comparison of different decouple structures on the MSCOCO validation set. We compare the performance of the decoder fully decoupled and split cross attention block only. }
\label{tab:table3}
\end{table}

\section{Conclusion}

Our work proposes a spatially decoupled DETR model that effectively addresses feature and prediction misalignment for both classification and box localization tasks. Specifically, we achieve this by splitting the cross-attention block in the decoder and enabling each branch to focus on its own region of interest while sharing the self-attention block. To further alleviate the misalignment of high classification confidence and precise localization we also introduce task alignment loss. Our experiments demonstrate the effectiveness of our approach. Future work will focus on exploring more complex transformer cross-attention structures to further decouple information.

\section{Acknowledgment}
This project is funded in part by the National Key R\&D Program of China Project 2022ZD0161100, by the Centre for Perceptual and Interactive Intelligence (CPII) Ltd under the Innovation and Technology Commission (ITC)’s InnoHK, by General Research Fund of Hong Kong RGC Project 14204021. Hongsheng Li is a PI of CPII under the InnoHK.
\newpage


{\small
\bibliographystyle{ieee_fullname}
\bibliography{main}

\begin{thebibliography}{10}\itemsep=-1pt

\bibitem{bochkovskiy2020yolov4}
Alexey Bochkovskiy, Chien-Yao Wang, and Hong-Yuan~Mark Liao.
\newblock Yolov4: Optimal speed and accuracy of object detection.
\newblock {\em arXiv preprint arXiv:2004.10934}, 2020.

\bibitem{cai2018cascade}
Zhaowei Cai and Nuno Vasconcelos.
\newblock Cascade r-cnn: Delving into high quality object detection.
\newblock In {\em Proceedings of the IEEE conference on computer vision and pattern recognition}, pages 6154--6162, 2018.

\bibitem{carion2020end}
Nicolas Carion, Francisco Massa, Gabriel Synnaeve, Nicolas Usunier, Alexander Kirillov, and Sergey Zagoruyko.
\newblock End-to-end object detection with transformers.
\newblock In {\em European conference on computer vision}, pages 213--229. Springer, 2020.

\bibitem{chen2022group}
Qiang Chen, Xiaokang Chen, Gang Zeng, and Jingdong Wang.
\newblock Group detr: Fast training convergence with decoupled one-to-many label assignment.
\newblock {\em arXiv preprint arXiv:2207.13085}, 2022.

\bibitem{duan2019centernet}
Kaiwen Duan, Song Bai, Lingxi Xie, Honggang Qi, Qingming Huang, and Qi Tian.
\newblock Centernet: Keypoint triplets for object detection.
\newblock In {\em Proceedings of the IEEE/CVF international conference on computer vision}, pages 6569--6578, 2019.

\bibitem{feng2021tood}
Chengjian Feng, Yujie Zhong, Yu Gao, Matthew~R Scott, and Weilin Huang.
\newblock Tood: Task-aligned one-stage object detection.
\newblock In {\em 2021 IEEE/CVF International Conference on Computer Vision (ICCV)}, pages 3490--3499. IEEE Computer Society, 2021.

\bibitem{gao2021fast}
Peng Gao, Minghang Zheng, Xiaogang Wang, Jifeng Dai, and Hongsheng Li.
\newblock Fast convergence of detr with spatially modulated co-attention.
\newblock In {\em Proceedings of the IEEE/CVF International Conference on Computer Vision}, pages 3621--3630, 2021.

\bibitem{gao2021mutual}
Ziteng Gao, Limin Wang, and Gangshan Wu.
\newblock Mutual supervision for dense object detection.
\newblock In {\em Proceedings of the IEEE/CVF International Conference on Computer Vision}, pages 3641--3650, 2021.

\bibitem{he2016deep}
Kaiming He, Xiangyu Zhang, Shaoqing Ren, and Jian Sun.
\newblock Deep residual learning for image recognition.
\newblock In {\em Proceedings of the IEEE conference on computer vision and pattern recognition}, pages 770--778, 2016.

\bibitem{he2022destr}
Liqiang He and Sinisa Todorovic.
\newblock Destr: Object detection with split transformer.
\newblock In {\em Proceedings of the IEEE/CVF Conference on Computer Vision and Pattern Recognition}, pages 9377--9386, 2022.

\bibitem{huang2022flowformer}
Zhaoyang Huang, Xiaoyu Shi, Chao Zhang, Qiang Wang, Ka~Chun Cheung, Hongwei Qin, Jifeng Dai, and Hongsheng Li.
\newblock Flowformer: A transformer architecture for optical flow.
\newblock In {\em European Conference on Computer Vision}, pages 668--685. Springer, 2022.

\bibitem{jia2022detrs}
Ding Jia, Yuhui Yuan, Haodi He, Xiaopei Wu, Haojun Yu, Weihong Lin, Lei Sun, Chao Zhang, and Han Hu.
\newblock Detrs with hybrid matching.
\newblock {\em arXiv preprint arXiv:2207.13080}, 2022.

\bibitem{jiang2018acquisition}
Borui Jiang, Ruixuan Luo, Jiayuan Mao, Tete Xiao, and Yuning Jiang.
\newblock Acquisition of localization confidence for accurate object detection.
\newblock In {\em Proceedings of the European conference on computer vision (ECCV)}, pages 784--799, 2018.

\bibitem{law2018cornernet}
Hei Law and Jia Deng.
\newblock Cornernet: Detecting objects as paired keypoints.
\newblock In {\em Proceedings of the European conference on computer vision (ECCV)}, pages 734--750, 2018.

\bibitem{li2022dn}
Feng Li, Hao Zhang, Shilong Liu, Jian Guo, Lionel~M Ni, and Lei Zhang.
\newblock Dn-detr: Accelerate detr training by introducing query denoising.
\newblock In {\em Proceedings of the IEEE/CVF Conference on Computer Vision and Pattern Recognition}, pages 13619--13627, 2022.

\bibitem{li2020generalized}
Xiang Li, Wenhai Wang, Lijun Wu, Shuo Chen, Xiaolin Hu, Jun Li, Jinhui Tang, and Jian Yang.
\newblock Generalized focal loss: Learning qualified and distributed bounding boxes for dense object detection.
\newblock {\em Advances in Neural Information Processing Systems}, 33:21002--21012, 2020.

\bibitem{li2019scale}
Yanghao Li, Yuntao Chen, Naiyan Wang, and Zhaoxiang Zhang.
\newblock Scale-aware trident networks for object detection.
\newblock In {\em Proceedings of the IEEE/CVF International Conference on Computer Vision}, pages 6054--6063, 2019.

\bibitem{lin2017feature}
Tsung-Yi Lin, Piotr Doll{\'a}r, Ross Girshick, Kaiming He, Bharath Hariharan, and Serge Belongie.
\newblock Feature pyramid networks for object detection.
\newblock In {\em Proceedings of the IEEE conference on computer vision and pattern recognition}, pages 2117--2125, 2017.

\bibitem{lin2017focal}
Tsung-Yi Lin, Priya Goyal, Ross Girshick, Kaiming He, and Piotr Doll{\'a}r.
\newblock Focal loss for dense object detection.
\newblock In {\em Proceedings of the IEEE international conference on computer vision}, pages 2980--2988, 2017.

\bibitem{lin2014microsoft}
Tsung-Yi Lin, Michael Maire, Serge Belongie, James Hays, Pietro Perona, Deva Ramanan, Piotr Doll{\'a}r, and C~Lawrence Zitnick.
\newblock Microsoft coco: Common objects in context.
\newblock In {\em European conference on computer vision}, pages 740--755. Springer, 2014.

\bibitem{liu2021switchable}
Boxiao Liu, Guanglu Song, Manyuan Zhang, Haihang You, and Yu Liu.
\newblock Switchable k-class hyperplanes for noise-robust representation learning.
\newblock In {\em Proceedings of the IEEE/CVF International Conference on Computer Vision}, pages 3019--3028, 2021.

\bibitem{liu2021fuseformer}
Rui Liu, Hanming Deng, Yangyi Huang, Xiaoyu Shi, Lewei Lu, Wenxiu Sun, Xiaogang Wang, Jifeng Dai, and Hongsheng Li.
\newblock Fuseformer: Fusing fine-grained information in transformers for video inpainting.
\newblock In {\em Proceedings of the IEEE/CVF international conference on computer vision}, pages 14040--14049, 2021.

\bibitem{liu2022dab}
Shilong Liu, Feng Li, Hao Zhang, Xiao Yang, Xianbiao Qi, Hang Su, Jun Zhu, and Lei Zhang.
\newblock Dab-detr: Dynamic anchor boxes are better queries for detr.
\newblock {\em arXiv preprint arXiv:2201.12329}, 2022.

\bibitem{liu2016ssd}
Wei Liu, Dragomir Anguelov, Dumitru Erhan, Christian Szegedy, Scott Reed, Cheng-Yang Fu, and Alexander~C Berg.
\newblock Ssd: Single shot multibox detector.
\newblock In {\em European conference on computer vision}, pages 21--37. Springer, 2016.

\bibitem{liu2019towards}
Yu Liu et~al.
\newblock Towards flops-constrained face recognition.
\newblock In {\em Proceedings of the IEEE/CVF International Conference on Computer Vision Workshops}, pages 0--0, 2019.

\bibitem{loshchilov2017decoupled}
Ilya Loshchilov and Frank Hutter.
\newblock Decoupled weight decay regularization.
\newblock {\em arXiv preprint arXiv:1711.05101}, 2017.

\bibitem{meng2021conditional}
Depu Meng, Xiaokang Chen, Zejia Fan, Gang Zeng, Houqiang Li, Yuhui Yuan, Lei Sun, and Jingdong Wang.
\newblock Conditional detr for fast training convergence.
\newblock In {\em Proceedings of the IEEE/CVF International Conference on Computer Vision}, pages 3651--3660, 2021.

\bibitem{redmon2016you}
Joseph Redmon, Santosh Divvala, Ross Girshick, and Ali Farhadi.
\newblock You only look once: Unified, real-time object detection.
\newblock In {\em Proceedings of the IEEE conference on computer vision and pattern recognition}, pages 779--788, 2016.

\bibitem{ren2015faster}
Shaoqing Ren, Kaiming He, Ross Girshick, and Jian Sun.
\newblock Faster r-cnn: Towards real-time object detection with region proposal networks.
\newblock {\em Advances in neural information processing systems}, 28, 2015.

\bibitem{shi2023videoflow}
Xiaoyu Shi, Zhaoyang Huang, Weikang Bian, Dasong Li, Manyuan Zhang, Ka~Chun Cheung, Simon See, Hongwei Qin, Jifeng Dai, and Hongsheng Li.
\newblock Videoflow: Exploiting temporal cues for multi-frame optical flow estimation.
\newblock {\em arXiv preprint arXiv:2303.08340}, 2023.

\bibitem{shi2023flowformer++}
Xiaoyu Shi, Zhaoyang Huang, Dasong Li, Manyuan Zhang, Ka~Chun Cheung, Simon See, Hongwei Qin, Jifeng Dai, and Hongsheng Li.
\newblock Flowformer++: Masked cost volume autoencoding for pretraining optical flow estimation.
\newblock In {\em Proceedings of the IEEE/CVF Conference on Computer Vision and Pattern Recognition}, pages 1599--1610, 2023.

\bibitem{song2020revisiting}
Guanglu Song, Yu Liu, and Xiaogang Wang.
\newblock Revisiting the sibling head in object detector.
\newblock In {\em Proceedings of the IEEE/CVF Conference on Computer Vision and Pattern Recognition}, pages 11563--11572, 2020.

\bibitem{wang2021pnp}
Tao Wang, Li Yuan, Yunpeng Chen, Jiashi Feng, and Shuicheng Yan.
\newblock Pnp-detr: Towards efficient visual analysis with transformers.
\newblock In {\em Proceedings of the IEEE/CVF International Conference on Computer Vision}, pages 4661--4670, 2021.

\bibitem{wang2022anchor}
Yingming Wang, Xiangyu Zhang, Tong Yang, and Jian Sun.
\newblock Anchor detr: Query design for transformer-based detector.
\newblock In {\em Proceedings of the AAAI conference on artificial intelligence}, volume~36, pages 2567--2575, 2022.

\bibitem{wu2020rethinking}
Yue Wu, Yinpeng Chen, Lu Yuan, Zicheng Liu, Lijuan Wang, Hongzhi Li, and Yun Fu.
\newblock Rethinking classification and localization for object detection.
\newblock In {\em Proceedings of the IEEE/CVF conference on computer vision and pattern recognition}, pages 10186--10195, 2020.

\bibitem{zhang2022accelerating}
Gongjie Zhang, Zhipeng Luo, Yingchen Yu, Kaiwen Cui, and Shijian Lu.
\newblock Accelerating detr convergence via semantic-aligned matching.
\newblock In {\em Proceedings of the IEEE/CVF Conference on Computer Vision and Pattern Recognition}, pages 949--958, 2022.

\bibitem{zhang2022dino}
Hao Zhang, Feng Li, Shilong Liu, Lei Zhang, Hang Su, Jun Zhu, Lionel~M Ni, and Heung-Yeung Shum.
\newblock Dino: Detr with improved denoising anchor boxes for end-to-end object detection.
\newblock {\em arXiv preprint arXiv:2203.03605}, 2022.

\bibitem{zhang2022towards}
Manyuan Zhang, Guanglu Song, Yu Liu, and Hongsheng Li.
\newblock Towards robust face recognition with comprehensive search.
\newblock In {\em European Conference on Computer Vision}, pages 720--736. Springer, 2022.

\bibitem{zhang2020discriminability}
Manyuan Zhang, Guanglu Song, Hang Zhou, and Yu Liu.
\newblock Discriminability distillation in group representation learning.
\newblock In {\em Computer Vision--ECCV 2020: 16th European Conference, Glasgow, UK, August 23--28, 2020, Proceedings, Part X 16}, pages 1--19. Springer, 2020.

\bibitem{zhu2019feature}
Chenchen Zhu, Yihui He, and Marios Savvides.
\newblock Feature selective anchor-free module for single-shot object detection.
\newblock In {\em Proceedings of the IEEE/CVF conference on computer vision and pattern recognition}, pages 840--849, 2019.

\bibitem{zhu2020deformable}
Xizhou Zhu, Weijie Su, Lewei Lu, Bin Li, Xiaogang Wang, and Jifeng Dai.
\newblock Deformable detr: Deformable transformers for end-to-end object detection.
\newblock {\em arXiv preprint arXiv:2010.04159}, 2020.

\end{thebibliography}
}

\end{document}